\documentclass[11pt,a4paper]{article}

\usepackage{amsmath,amsfonts,bm}

\def\eqref#1{equation~\ref{#1}}

\def\1{\bm{1}}

\def\rva{{\mathbf{a}}}

\def\rvz{{\mathbf{z}}}

\def\rmA{{\mathbf{A}}}

\def\rmX{{\mathbf{X}}}
\def\rmY{{\mathbf{Y}}}
\def\rmZ{{\mathbf{Z}}}

\def\mH{{\bm{H}}}

\def\mK{{\bm{K}}}

\def\mQ{{\bm{Q}}}

\def\mV{{\bm{V}}}
\def\mW{{\bm{W}}}

\DeclareMathAlphabet{\mathsfit}{\encodingdefault}{\sfdefault}{m}{sl}
\SetMathAlphabet{\mathsfit}{bold}{\encodingdefault}{\sfdefault}{bx}{n}

\newcommand{\softmax}{\mathrm{softmax}}

\usepackage[hyperref]{acl2021}
\usepackage{times}
\usepackage{latexsym}
\usepackage{tikz}
\usepackage{booktabs}
\usepackage{hyperref}
\usepackage{multirow}
\usepackage{subcaption}
\usepackage{adjustbox}
\usepackage{xspace}
\usetikzlibrary{fit,positioning,decorations.pathmorphing,calc}
\usepackage[T1]{fontenc}

\usepackage{bm}
\usepackage{wrapfig}
\usepackage{changepage}
\usepackage{amsmath}
\usepackage{amsthm}
\usepackage{amssymb}
\usepackage{amsfonts,eucal,amsbsy}
\usepackage{mathtools}
\usepackage{multirow}
\usepackage[normalem]{ulem}
\usepackage{natbib}

\usepackage{booktabs}
\usepackage{tablefootnote}

\newcommand\numberthis{\addtocounter{equation}{1}\tag{\theequation}}

\newenvironment{itemizesquish}{\begin{list}{\labelitemi}{\setlength{\itemsep}{-0.4em}\setlength{\labelwidth}{0.5em}\setlength{\leftmargin}{\labelwidth}\addtolength{\leftmargin}{\labelsep}}}{\end{list}}

\newcommand{\model}{\textsc{coda}\xspace}
\newcommand{\base}{\textsc{transformer}\xspace}
\newcommand\myDots{\makebox[1em][c]{.\hfil.\hfil.}}
\newcommand{\wmt}{\texttt{WMT14 EN-DE}\xspace}
\newcommand{\iwslt}{\texttt{IWSLT14 DE-EN}\xspace}
\usepackage{microtype}

\aclfinalcopy 
\def\aclpaperid{2170} 

\title{Cascaded Head-colliding Attention}

\author{%
Lin Zheng$^{\spadesuit\diamondsuit}$ ~ Zhiyong Wu$^{\spadesuit}$ ~ Lingpeng Kong$^{\spadesuit\clubsuit}$\\
$^{\spadesuit}$Department of Computer Science, The University of Hong Kong\\
$^{\diamondsuit}$School of Data and Computer Science, Sun Yat-sen University\\
$^\clubsuit$Shanghai Artificial Intelligence Laboratory\\
\tt zhenglin6@mail2.sysu.edu.cn,\{zywu,lpk\}@cs.hku.hk\\
}

\date{}

\begin{document}
\maketitle
\begin{abstract}
Transformers have advanced the field of natural language processing (NLP) in many ways. At the heart of the Transformer architecture is the multi-head attention (MHA) mechanism which models pairwise interactions between the elements of the sequence. Despite its massive success, the current framework ignores interactions among different heads, leading to the problem that many of the heads are redundant in practice, which underutilizes the capacity of the model. To improve parameter efficiency, we re-formulate the MHA as a latent variable model from a probabilistic perspective. We present \textbf{c}ascaded head-c\textbf{o}lli\textbf{d}ing \textbf{a}ttention (\model) which explicitly models the interactions between attention heads through a hierarchical variational distribution. We conduct extensive experiments and demonstrate that \model outperforms the transformer baseline, by $0.6$ perplexity on \texttt{Wikitext-103} in language modeling, and by $0.6$ BLEU on \texttt{WMT14 EN-DE} in machine translation, due to its improvements on the parameter efficiency.\footnote{Our implementation is publicly available at \url{https://github.com/LZhengisme/CODA}.}
\end{abstract}

\section{Introduction}
\label{sec:intro}
Transformers \citep{vaswani2017attention} have advanced the field of natural language processing (NLP) on a variety of important tasks, including language modeling \citep{dai-etal-2019-transformer,baevski2018adaptive}, language understanding \citep{devlin-etal-2019-bert,yang2019xlnet}, and machine translation \citep{vaswani2017attention,dehghani2018universal,liu2020very}. It has also found its place in computer vision \citep{dosovitskiy2020image}, and in intelligent agents \citep{vinyals2019grandmaster} where sequence modeling plays a key role as well. The cornerstone of the transformer architecture is the multi-head attention (MHA) mechanism which models pairwise interactions between the elements of the sequence. An attention function can be described as mapping a query and a set of key-value pairs to an output, where the query, keys, values, and output are all vectors. The output is computed as a weighted sum of the values, where the weight assigned to each value is computed by a compatibility function of the query with the corresponding key. A multi-head attention (MHA) mechanism extends the idea through performing multiple separately parameterized attention functions acting in parallel to contextualize the input representations. Their outputs are then gathered by an affine transformation, allowing the model to jointly attend to information from different representation subspaces at different positions. 

Despite its massive success, the current framework ignores the interactions among different heads, leading to the problem that many of the heads are redundant in practice (i.e., attending to the same regions of the sequence), which underutilizes the capacity of the model \citep{voita-etal-2019-analyzing,michel2019sixteen}. At the same time, recent research \citep[\emph{inter alia}]{tang-etal-2018-self,clark2019does,voita-etal-2019-analyzing,wu-etal-2020-perturbed} demonstrates that heads in MHA have the potential to capture distinct information from input sequences, ranging from syntactic and semantic features to alignment information between source and target sentence pairs. These observations suggest that multiple heads should be encouraged to extract complementary information. Therefore, it is highly appealing to take into account the interactions among different attention heads from the perspective of parameter efficiency and the expressiveness of the model.

In this work, we introduce \textit{head-colliding attention} (\S\ref{sec:model}). We formulate MHA as a probabilistic model, where each attention head is represented by a latent variable and all of them \textit{collide} into the observed sequence data  (Figure~\ref{fig:hca}). In this probabilistic graphical model structure, attention heads work as individual factors to explain the data. Although each factor is independent of each other \textit{a priori}, they interact with each other automatically, conditioning on observations, thanks to the explaining-away effects \citep{Pearl1989ProbabilisticRI,Wellman1993ExplainingA}.


The head-colliding attention mechanism introduces new computational challenges in training the model. We will discuss how we tackle these using variational methods \citep{blei2017variational}. We propose \textbf{c}ascaded head-c\textbf{o}lli\textbf{d}ing \textbf{a}ttention (\model, Figure~\ref{fig:coda}). As our main model, \model adopts a hierarchical variational distribution \citep{ranganath2016hierarchical} to allow both rich head interactions and effective computations (\S\ref{sec:training}).



We validate our method in language modeling and machine translation experiments (\S\ref{sec:ex-analysis}). \model outperforms the vanilla MHA transformer on both tasks, on \texttt{Wikitext-103} by $0.6$ perplexity and on \texttt{WMT14 EN-DE} by $0.6$ BLEU. Further analysis shows that \model learns to encourage diversity in different heads (Figure~\ref{fig:jsd-wikitext-8heads}) and to promote parameter efficiency when increasing the number of heads (\S\ref{ssec:ex-num-heads}).



\section{Background}
\label{sec:bg}


Multi-head attention (MHA) mechanism plays an important role in modern transformer architecture \citep{vaswani2017attention}. It extends the classical attention mechanism by running multiple attention function heads in parallel.

An MHA module is composed of $h$ identical blocks (usually referred to as attention heads). Each head will generate a hidden state $\mH_i$ based on the input Query, Key and Value matrices, denoted as $\mQ$, $\mK$, and $\mV$ respectively. The hidden states from different heads are then aggregated as the output of the MHA module: $\sum_{i=1}^n \mH_i \mW_i^o$, where $\mW_i^o$ are model parameters.

In the $i$-th head, the input matrices $\mQ$, $\mK$ and $\mV$ are first linearly projected into different subspace representations $\widetilde{\mQ}_i$, $\widetilde{\mK}_i$, and $\widetilde{\mV}_i$, based on different learnable parameters. After that, we compute the inner product over all projected queries and keys as the attention logits $\rvz_i$, which are then passed through a row-wise softmax\footnote{We omit the scaling factor for simplicity.} to obtain head attention weights $\rva_i$:
\begin{equation}
\label{eqn:z-formula}
     \rva_i = \softmax(\rvz_i) = \softmax(\widetilde{\mQ}_i\widetilde{\mK}_i^T).
\end{equation}

 
The final output of a single attention block is the weighted sum of $\widetilde{\mV}_i$:
\begin{align*}
  \mH_i = \rva_i \widetilde{\mV}_i.
\end{align*}
As we can see, the core of MHA is to calculate $\rva_i$ in each head. We thus refer to $\rva_i$ as the $i$-th attention head.


In sequence prediction tasks, the model takes as input a source sequence of length $m$ and outputs a target sequence of length $n$ in an auto-regressive manner. It predicts each token $\rmY$ within the target sequence through a categorical distribution $p_{\mathrm{vanilla}}(\rmY|\rmX)$, where $\rmX$ includes the source sequence as well as a previously generated prefix. With respect to an MHA block $\rva_1, \dots, \rva_h$, the model predicts target tokens $\rmY$ by first feeding these heads into a complex non-linear transformation\footnote{Since a transformer typically stacks several attentive layers, for an MHA block in some layer, subsequent layers will induce a non-linear transformation $\phi(\cdot)$ for its attention heads. For instance, $\phi(\cdot)$ may include several other MHA blocks and feed-forward networks.} denoted by $\phi(\cdot)$, and then passing it through a $\mathrm{softmax}$ function over the entire vocabulary. Therefore, the output probability can be written as $p_{\mathrm{vanilla}}(\rmY|\rmX) = f(\rva_1, \dots, \rva_h)$, where
\begin{align*}
    f(\rva_1, \dots, \rva_h) \coloneqq \softmax(\phi(\rva_1, \dots, \rva_h)).
\end{align*}


\section{Head-colliding Attention}
\label{sec:model}
\begin{figure}
\centering
\begin{subfigure}[t]{0.45\columnwidth} 
  \centering
  	\adjustbox{max totalsize={\columnwidth}{\textheight},center}{    
    \begin{tikzpicture}
  \tikzstyle{main}=[circle, minimum size = 9mm, thick, draw =black!80, node distance = 5mm]
  \tikzstyle{connect}=[-latex, thick]
  \tikzstyle{box}=[rectangle, draw=black!100]
    \node[main, fill = white!100] (a1)  { $\rva_1$ };
    \node[main, fill = white!100] (a2) [right=of a1] { $\rva_2$ };
    \node[main, fill = white!100] (ah) [right=of a2, xshift=10mm] { $\rva_h$ };
    \node[main, fill = black!10] (y) [below=of a1, xshift=20mm] { $\rmY$ };
    \path (a1) edge [connect] (y)
          (a2) edge [connect] (y)
      (ah) edge [connect] (y)
      (a2) -- node[auto=false]{\ldots} (ah);
        \node[below=1.4cm of y] (dummy){};
  \end{tikzpicture}
  }
  \caption{Head-colliding attention.}
  \label{fig:hca}
\end{subfigure}\quad
\begin{subfigure}[t]{0.45\columnwidth} 
  \centering
  \adjustbox{max totalsize={\columnwidth}{\textheight},center}{
      \begin{tikzpicture}
  \tikzstyle{main}=[circle, minimum size = 9mm, thick, draw =black!80, node distance = 5mm]
  \tikzstyle{every pin edge}=[<-,shorten <=1pt]
  \tikzstyle{connect}=[-latex]
  \tikzstyle{box}=[rectangle, draw=black!100]
    \node[main, fill = white!100] (a11)  { $\rva_1^1$ };
    \node[main, fill = white!100] (a12) [right=of a11] { $\rva_2^1$ };
    \node[main, fill = white!100] (a13) [right=of a12, xshift=10mm] { $\rva_h^1$ };
    
    \node[main, fill = white!100] (a21) [below=of a11, yshift=-5mm] { $\rva_1^2$ };
    \node[main, fill = white!100] (a22) [right=of a21] { $\rva_2^2$ };
    \node[main, fill = white!100] (a23) [right=of a22, xshift=10mm] { $\rva_h^2$ };
    
    \node[main, fill = white!100] (a31) [below=of a21, yshift=-5mm] { $\rva_1^L$ };
    \node[main, fill = white!100] (a32) [right=of a31] { $\rva_2^L$ };
    \node[main, fill = white!100] (a33) [right=of a32, xshift=10mm] { $\rva_h^L$ };
    \node[main, fill = black!10] (y) [below=of a31, xshift=20mm] { $\rmY$ };
    \foreach \i in {1,...,3}
        \path (a\i 2) -- node[auto=false]{\ldots} (a\i 3);
    
    \foreach \j in {1,...,3}
        \foreach \i in {1,...,3}
        {
            \path (a2\i) edge [connect] (a3\j);
            \path (a1\i) edge [connect] (a2\j);
        }
    
    \path (a31) edge [connect] (y)
        (a32) edge [connect] (y)
        (a33) edge [connect] (y);
    
  \end{tikzpicture}
  }
  
  \caption{Cascaded head-colliding attention (\model).}
  \label{fig:coda}
\end{subfigure}
\caption{(a) Left: Probabilistic graphical model (PGM) diagram of head-colliding attention. Although each head variable is independent \textit{a priori}, they interact with each other \textit{after} observing targets $\rmY$, which is referred as explaining-away effect. (b) Right: PGM diagram of a 3-layer cascaded head-colliding attention (\model). $\rva_{i}^{l}$ denotes the $i$-th attention head at transformer layer $l$. Note that all dependencies from $\rmX$ are omitted in these diagrams for simplicity.}
\label{fig:pgm}
\end{figure}
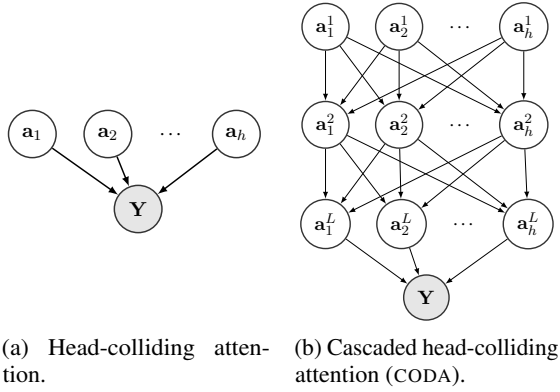

In this section, we introduce \textit{head-colliding attention}. Specifically, we formulate MHA as a probabilistic model, where each attention head is represented by a latent variable. The name reflects a ``\textit{collider}'' in the context of probabilistic graphical models (Figure~\ref{fig:hca}). We will first explain how head-colliding attention permits the modeling of interactions among different heads and then discuss how vanilla MHA can be viewed as a marginalized version of head-colliding attention, which ignores any head interactions.

Considering a single MHA block, we cast each attention head $\rva_i$ as a latent variable. The probability of target $\rmY$ conditioning on input $\rmX$ can be obtained by marginalizing over all heads $\rmA$ (we denote $\rmA \coloneqq \{\rva_1, \dots,\rva_h\}$):
\begin{align*}
p(\rmY|\rmX)
&= \int_{\rmA} p(\rmY|\rmA, \rmX) p(\rmA | \rmX) d \rmA \\
&= \mathbb{E}_{p(\rmA|\rmX)} \left[f(\rmA)\right].
\end{align*}
$p(\rmA | \rmX)$ is the joint prior distribution. The corresponding directed graphical model is demonstrated in Figure~\ref{fig:hca}, where the links from different heads collide on the observation variable $\rmY$. A crucial property of this graphical model is the ``explaining-away'' effect \citep{Pearl1989ProbabilisticRI,Wellman1993ExplainingA} of attention heads $\rmA$ when observing the output $\rmY$. In other words, if a head $\rva_i$ attends to part of the input which accords well with observation, it immediately discourages other heads from attending to the same part of the input but encourages them to look into complementary information.\footnote{In other words, if we confirm that some head accords well with the observation, then the probability of other heads should be reduced since there is less need to invoke them, according to Occam’s razor.} This mechanism effectively reduces head redundancy and in turn improves parameter efficiency.

  
\paragraph{Vanilla vs. head-colliding attention} We now take a closer look at the vanilla MHA (\S\ref{sec:bg}). 
Recall that in vanilla MHA, all attention heads are deterministic. From the perspective of latent variable models, this is computationally equivalent to taking expectations of latent head variables. The output probability distribution $p_{\mathrm{vanilla}}(\rmY|\rmX)$ can then be expressed as:
\begin{align}
f(\mathbb{E}_{p(\rva_1|\rmX)}\left[\rva_1\right], \dots, \mathbb{E}_{p(\rva_h|\rmX)}\left[\rva_h\right]).
    \label{eqn:vanilla-approx}
\end{align}
This means we are only interested in the individual expectations when using the attention heads in vanilla MHA for predictions.
On the contrary, in head-colliding attention the distribution of $\rmY$ is defined as:
\begin{align*}
  p(\rmY|\rmX)
  &= \mathbb{E}_{p(\rva_1, \dots, \rva_h|\rmX)}  \left[ f(\rva_1, \dots, \rva_h)\right].
\end{align*}
Note the inherent difference of when to take the expectation in vanilla and head-colliding attention.
Since $f(\cdot)$ is a complex non-linear function (\S\ref{sec:bg}), these two formulations are not equivalent in general and may have a large gap between the two distributions. Concretely, vanilla MHA ignores any possible interactions among different heads. As indicated in \eqref{eqn:vanilla-approx}, it first marginalizes out every single head \textit{before} observing targets -- one head will not learn what other heads are attending to despite the fact $\rmY$ is observed. This is why vanilla MHA is prone to redundancy as many previous studies \citep[\emph{inter alia}]{voita-etal-2019-analyzing,michel2019sixteen} discovered.
Head-colliding attention, on the other hand, permits rich head interactions due to the expressive non-linear function $f(\cdot)$ inside the expectation over different latent variables $\rva_1, \dots, \rva_h$. However, the complexity of head interactions also leads to intractability in training the model, which we will discuss in the next section. 


\section{Training Head-colliding Attention}
\label{sec:training}
We train the model by performing maximum likelihood estimation. Here, the log marginal likelihood can be expressed as:
\begin{align*}
\log p(\rmY|\rmX) = \log \mathbb{E}_{p(\rmA | \rmX)} \left[ p(\rmY|\rmA, \rmX)\right].
\end{align*}
Unfortunately, this is intractable in general because it requires marginalizing over all possible configurations of attention heads. The standard technique is to use variational inference, which optimizes the log marginal by maximizing its evidence lower bound (called ELBO) \citep{blei2017variational}:
\begin{align*}
\mathcal{L} &\coloneqq \mathbb{E}_{q(\rmA | \rmX)} \left[\log\frac{p(\rmY | \rmA, \rmX)p(\rmA | \rmX)}{q(\rmA|\rmX)}\right] \numberthis{}\label{eqn:obj}\\
&= \log p(\rmY|\rmX) - \mathrm{KL}(q(\rmA | \rmX) || p(\rmA | \rmX, \rmY)) \\
&\leq \log p(\rmY|\rmX),
\end{align*}
where $q(\rmA|\rmX)$ is the variational distribution\footnote{Although the variational distribution $q$ should depend on target $\rmY$ in principle, such conditioning renders testing difficult since the target information is not available during testing. For this reason, we only consider the source $\rmX$ hereafter.} over latent variables $\rmA$. $p(\rmA | \rmX, \rmY)$ is the intractable posterior distribution of all heads given observations $\rmY$ and the input $\rmX$, which encodes the rich head interactions we desire, as discussed in \S\ref{sec:model}.
Therefore, an ideal variational distribution $q(\rmA|\rmX)$ should be close to the true posterior $p(\rmA | \rmX, \rmY)$. In this case, the samples would accurately reflect the head interactions and the variational distribution would yield a tighter bound to $\mathcal{L}$ to facilitate the training. 

A straight-forward choice of $q(\rmA|\rmX)$ is to use the mean-field approximation~\citep{vae1}:
\begin{align*}
  q(\rmA|\rmX) = q(\rva_1, \rva_2, \dots, \rva_h |\rmX) = \prod_{i=1}^h q(\rva_i |\rmX).
\end{align*}
However, it has similar drawbacks as the vanilla MHA.\footnote{Note that the vanilla MHA does not define distributions over heads in its original context. We derive this from the latent-variable perspective.} The mean-field approximation assumes the independence of different heads and hence the interactions are greatly limited.

Alternatively, one could parameterize $q(\rmA|\rmX)$ using an auto-regressive model.\footnote{This works well in our preliminary experience, despite its extremely expensive computational cost.} Although this is much more expressive, its sequential nature severely slows down training, making it infeasible in practice.

\paragraph{Cascaded Head-colliding attention} Our solution to this problem is to employ hierarchical structures for head-colliding attention, where interactions among heads could be effectively incorporated into the model \citep{snderby2016ladder, ranganath2016hierarchical}. 

Conveniently, the hierarchical nature of the transformer architecture offers an effective way of constructing such proposal distributions. Given a transformer with $L$ layers, we denote the set of all attention heads at layer $l-1$ and $l$ as $\rmA^{l-1}$ and $\rmA^{l}$, respectively. Following the bottom-up computation of the transformer, the distribution of $\rmA^{l}$ must rely on the instantiated values of $\rmA^{l-1}$. In this sense, $\rmA^{l-1}$ can be seen as the common variables that govern $\rmA^{l}$ (Figure~\ref{fig:coda}). Formally, we have:
\begin{equation*}
    q(\rmA^{1}, \myDots, \rmA^{L} | \rmX) \!\!=\!\! q(\rmA^{1}|\rmX)\prod_{j = 2}^{L}q(\rmA^{j} | \rmX,\rmA^{j-1}).
\end{equation*}
Despite the fact that each attention head $\rva_i^l \in \rmA^{l}$ at $l$-th layer is conditionally independent given $\rmA^{l-1}$, they become dependent when we marginalize $\rmA^{l-1}$ out. In particular, the marginal distribution of each $\rmA^{l}$ becomes:
\begin{align*}
q(\rmA^{l} | \rmX) \!\! = \!\!\!\int_{\rmA^{l-1}} \!\!\! q(\rmA^{l-1}|\rmX) q(\rmA^{l} |\rmX,\rmA^{l-1}) d\rmA^{l-1}.
\end{align*}
This corresponds to an \textit{infinite mixture} of the mean-field distributions $q(\rmA^{l} |\rmX,\rmA^{l-1})$ and is able to capture rich head interactions \citep{ranganath2016hierarchical}. 
Our main model adopts this cascaded proposal distribution in figure~\ref{fig:coda}, and therefore we name it \textbf{c}ascaded head-c\textbf{o}lli\textbf{d}ing \textbf{a}ttention (\model).

The only problem left now is how to specify the conditional distribution $q(\rmA^{l}|\rmX,\rmA^{l-1})$ for all $l = 1,2,\dots,L$. We first impose the basic constraints on head values as in vanilla MHA, that is, all head values must range within a simplex $\Delta^{n-1}$:
\begin{align*}
  \Delta^{n-1} = \{\rmA^{l} | \sum_{k=1}^{n} \rva_{i,:k}^{l} = \mathbf{1}, \forall i=1,\dots,h\}.
\end{align*}
Here $\rva_{i,:k}^{l}$ is the $k$-th column of the $i$-th attention head at layer $l$ and $\mathbf{1}$ denotes the vector of all 1's. For efficient training and inference, we adopt Gaussian-logistic distributions \citep{blei2006correlated,cohen2008logistic}, which not only satisfy the constraints above but also benefit from the effective reparameterization trick \citep{vae1,vae2,vae3}. 

In particular, recall that in vanilla MHA, $\rva_i = \softmax(\rvz_i) = \softmax(\widetilde{\mQ}_i\widetilde{\mK}_i^{T})$ (\eqref{eqn:z-formula}). We also denote the attention logits at $l$-th layer as $\rmZ^{l} \coloneqq \{\rvz_1^{l},\dots,\rvz_h^{l}\}$. For head $i$ at layer $l$, we first sample from a multivariate Gaussian distribution $q(\rvz_{i,j:}^l | \rvz_{i,j:}^{l-1})$ \footnote{We only explicitly define the attention logit $\rvz$ as random variables, while the distribution of heads $\rva$ is induced via a deterministic transformation (\textit{i.e.}, softmax function) of $\rvz$; therefore it suffices to build dependencies between attentive logits instead.} and pass the samples into a row-wise softmax function to yield head values:
\begin{align*}
\rvz_{i,j:}^l \sim  \mathcal{N}(\mu_{i,j:}^l, \Sigma), \quad \rva_{i,j:}^l=\softmax(\rvz_{i,j:}^l),
\end{align*}
where $\rvz_{i,j:}^l$ and $\rva_{i,j:}^l$ represent the $j$-th row of the $i$-th attention logit and attention head at layer $l$ respectively.


To explicitly model hierarchical structures among attention heads, we propose to add a direct connection between attention heads at adjacent layers (Figure~\ref{fig:coda}). Such connections offer direct access to the information of attention in the previous layer.
Specifically, for each head $i$ at layer $l$ we set the  mean $\mathbf{\mu_i}^l$ as the sum of two parts:
\begin{align*}
\mathbf{\mu_i}^l = 
{\underbrace{\widetilde{\mQ}_i\widetilde{\mK}_i^T}_\text{vanilla MHA}
}\,+{
\underbrace{\sigma_i(\rmZ^{l-1})}_\text{direct connection}\!\!\!\!, \numberthis{}\label{eqn:direct-connection}
}
\end{align*}
where $\sigma_i(\cdot)$ is a two-layer multilayer perceptron (MLP) to fuse information from different heads $\rmZ^{l-1}$ (see the cascading connections in Figure~\ref{fig:coda} for an illustration). We set the covariance matrix $\Sigma$ to the identity matrix for all attentive logits. 
We give the prior the same form as the variational posterior and parameters are shared between $q(\rmA^{1}, \myDots, \rmA^{L} | \rmX)$ and $p(\rmA^{1}, \myDots, \rmA^{L} | \rmX)$ for our objective (\eqref{eqn:obj}). With the help of parameter sharing, the KL term in \eqref{eqn:obj} is also cancelled out due to the identical distributions.\footnote{Therefore, it can also be derived by directly applying the Jensen's inequality on the log marginal likelihood.} This choice works well in practice, where it not only allows \model to use almost the same amount of parameters as vanilla Transformer, but also eliminates the need to invoke advanced training techniques for amortized variational inference.\footnote{For instance, training a standard variational auto-encoder (VAE) for NLP tasks often suffers from the posterior collapse problem due to the heavy KL regularization \citep{bowman-etal-2016-generating}, where some tricks have to be used to achieve good performance, such as KL annealing, etc.} More details can be found in Appendix~\ref{app:implementation}.

\section{Experiments}
\label{sec:ex-analysis}
We conduct experiments on language modeling and machine translation tasks.
\subsection{Setup}
\label{ssec:setup}
\paragraph{Datasets}
First, we conducted experiments for token-level language modeling on a large-scale benchmark dataset \texttt{Wikitext-103} \citep{merity2016pointer}, which consists of articles from Wikipedia with the token number around 103M/218K/246K for the training/validation/testing splits respectively. The vocabulary size is 267,744.

For machine translation, we consider two standard datasets:
\begin{itemizesquish}
\item \wmt \citep{bojar2014findings}, which contains about 4.5M/3K/3K sentences pairs for training/validation/testing splits respectively. We follow \citet{ott-etal-2018-scaling} and \citet{peng-etal-2020-mixture} to preprocess the dataset, and obtain a shared vocabulary between source and target language of around 32K byte pair encoding (BPE, \citet{sennrich-etal-2016-neural}) types.
\item \iwslt \citep{cettolo2014report}. Following standard practice~\cite{edunov-etal-2018-classical,peng-etal-2020-mixture}, we pre-process the 160K/7K/7K sentence pairs and build training/validation/testing sets accordingly. This generates a vocabulary of around 9K(7K) BPE types for source(target).
\end{itemizesquish}

\paragraph{Implementation details}
We implement our model with PyTorch \citep{pytorch} and FairSeq toolkit \citep{ott-etal-2019-fairseq}. In particular, our model is based on the vanilla transformer architecture \citep{vaswani2017attention}. For \model, we replace all vanilla MHA blocks with the cascaded head-colliding attention, for both self attention and cross attention (if any). In language modeling, we use adaptive input embeddings \cite{baevski2018adaptive} and set context size to 512 and 480 for training and testing respectively, due to constraints of computational resources.
In machine translation, we set beam size to 5 and adopt the hyperparameters from \citep{peng-etal-2020-mixture} for \iwslt. For \wmt we set beam size to 4, length penalty to 0.6, and average last 10 checkpoints for testing, following \citet{vaswani2017attention}. Further implementation details can be found in Appendix~\ref{app:implementation}.

\subsection{Main results}
\label{ssec:ex-main}
The results of language modeling on \texttt{Wikitext-103} dataset are reported in Table~\ref{tb:wikitext103}. As we can see from the table, \model barely introduces any additional parameters. However, by taking into account head interactions, \model significantly outperforms \base by over 0.6 perplexity. For reference, we also report the best setting (denoted by  \base\!\!\textsuperscript{\textdagger}) in \citet{baevski2018adaptive}, which uses a much larger context size (3072/2560 \textit{vs.} 512/480 for training/testing), \model still outperforms by a substantial margin of 0.3 perplexity. This indicates that encouraging head interactions can improve parameter efficiency.


To show whether \model has promoted head interactions and reduced head redundancy, we qualitatively visualize the attention heads in both \model and \base via heatmaps. Concretely, we compute the Jensen-Shannon Divergence (JSD) between each pair of attention heads at the same layer.

In particular, we assume head values define a categorical distribution in both \base and \model model to facilitate comparison. That is, an attentive head $\rva_i$ induces $n$ categorical distributions for each query position. For the $j$-th distribution, it indicates how the $j$-th target position attends to all $m$ source positions and is denoted by $p(x|\rva_{i,j:})$. For two heads $i$ and $i'$, we first compute their average distribution as
\begin{align*}
    m &\coloneqq \frac{p(x|\rva_{i,j:}) + p(x|\rva_{i',j:})}{2}
\end{align*}
Then the $\mathrm{JSD}$ value between the $i$-th and $i'$-th attention head is computed by summing all of $n$ induced distributions:
\begin{align*}
    \sum_{j=1}^n \frac{1}{2}\left(\mathrm{KL}(p(x|\rva_{i,j:})||m)\! + \! \mathrm{KL}(p(x|\rva_{i',j:})||m))\right)
\end{align*}
We average computed JSDs for all validation samples. Note that a larger JSD value (darker color) indicates that two heads are behaving more differently (i.e. less redundancy between them), and vice versa.

\begin{figure*}
\centering
\includegraphics[width=\textwidth]{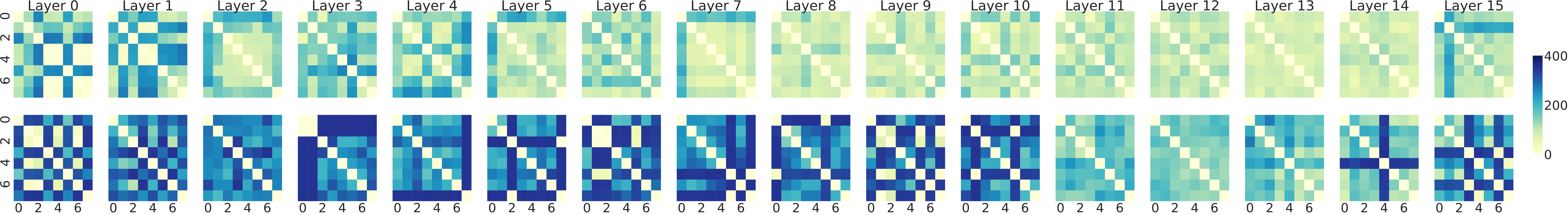}
\caption{Jensen-Shannon Divergences (JSD) for each pair of attention heads at all 16 layers on \texttt{Wikitext-103} validation dataset. Top: JSD heatmap of attention heads from \base model; Bottom: JSD heatmap of attention heads from \model. Columns represent different layers of both models. The darker color implies a larger divergence between two heads and in turn less redundancy.}
\label{fig:jsd-wikitext-8heads}
\end{figure*}

As shown in Figure~\ref{fig:jsd-wikitext-8heads}, JSD heatmaps in \model are clearly darker than those in \base. This suggests that \model permits richer head interactions, which fosters different heads to communicate with each other and encourages them to become complementary. Consequently, our model effectively reduces head redundancy in MHA and improves parameter-efficiency.

\begin{table}
	\centering
	\resizebox{0.9\columnwidth}{!}{    
		\begin{tabular}{c||c|c|c}
			\toprule
			{Model}   & {\# Params.} & {Val. PPL} & {Test PPL}\\
			\hline
            \base     & 246.93M &  18.35  &  19.08\\
            \base\!\textsuperscript{\textdagger}    &  246.93M    & 17.97 & 18.70 \\
			\hline
			\model     &  246.96M & \textbf{17.81} &  \textbf{18.48} \\   
			\bottomrule
	\end{tabular}}
	\caption{Validation (Val.) and testing Perplexity (PPL) on \texttt{Wikitext-103} dataset (lower is better). \base is the base model in \citet{baevski2018adaptive} with the same context size as \model (512/480 for training/testing), while \base\!\!\textsuperscript{\textdagger} is the same model but with the best setting in their paper, which uses much larger context size (3072/2560 respectively); the result for \base\!\!\textsuperscript{\textdagger} is as reported in \citet{baevski2018adaptive}. 
	}
	\label{tb:wikitext103}
\end{table}

The results on \iwslt and \wmt datasets are shown in Table~\ref{tb:mt-result}. We see that \model exhibits clear improvements over \base: a 1.1 point gain in BLEU on \iwslt dataset and a 0.6 BLEU improvement on \wmt dataset. Despite such significant gains over the baseline, \model only introduce very few additional parameters (\textit{e.g.}, 0.03\% extra parameters on \iwslt). This, again, shows that \model is more parameter efficient than vanilla Transformer due to the cascaded head-colliding attention we proposed.
Similar to experiments on language modeling, we also visualize the head behaviors to measure attentive head interactions (See Figure~\ref{fig:jsd-iwslt-4heads} and Figure~\ref{fig:jsd-wmt-8heads} in Appendix~\ref{app:figs}), where we observe similar phenomena on translation tasks. Specifically, different heads in \model are often complementary to each other and focus on quite different regions of sequences, rather than becoming redundant or even identical as observed in \base models.


\begin{table}
	\centering
	\resizebox{\columnwidth}{!}{    
		\begin{tabular}{c |c c || c c}
			\toprule
			\multirow{2}{*}{Model} & \multicolumn{2}{c||}{\iwslt} & \multicolumn{2}{c}{\wmt} \\
			\cline{2-5}
               & \# Params. & BLEU    & \# Params.    &  BLEU     \\
			\hline
			\base  & 39.47M  & 34.5  & 60.92M & 27.4\\
			\hline
			\model  & 39.48M  & \textbf{35.6}  & 60.94M & \textbf{28.0}\\
			\bottomrule
	\end{tabular}}  
	\caption{Performance of \base and \model on \iwslt and \wmt datasets.}
	\label{tb:mt-result}
\end{table}


\subsection{Analysis: the effect of the number of attention heads}
\label{ssec:ex-num-heads}
\begin{figure}[t]
      \centering
  	\begin{subfigure}[b]{.49\columnwidth}
      \centering
      {\includegraphics[width=\columnwidth]{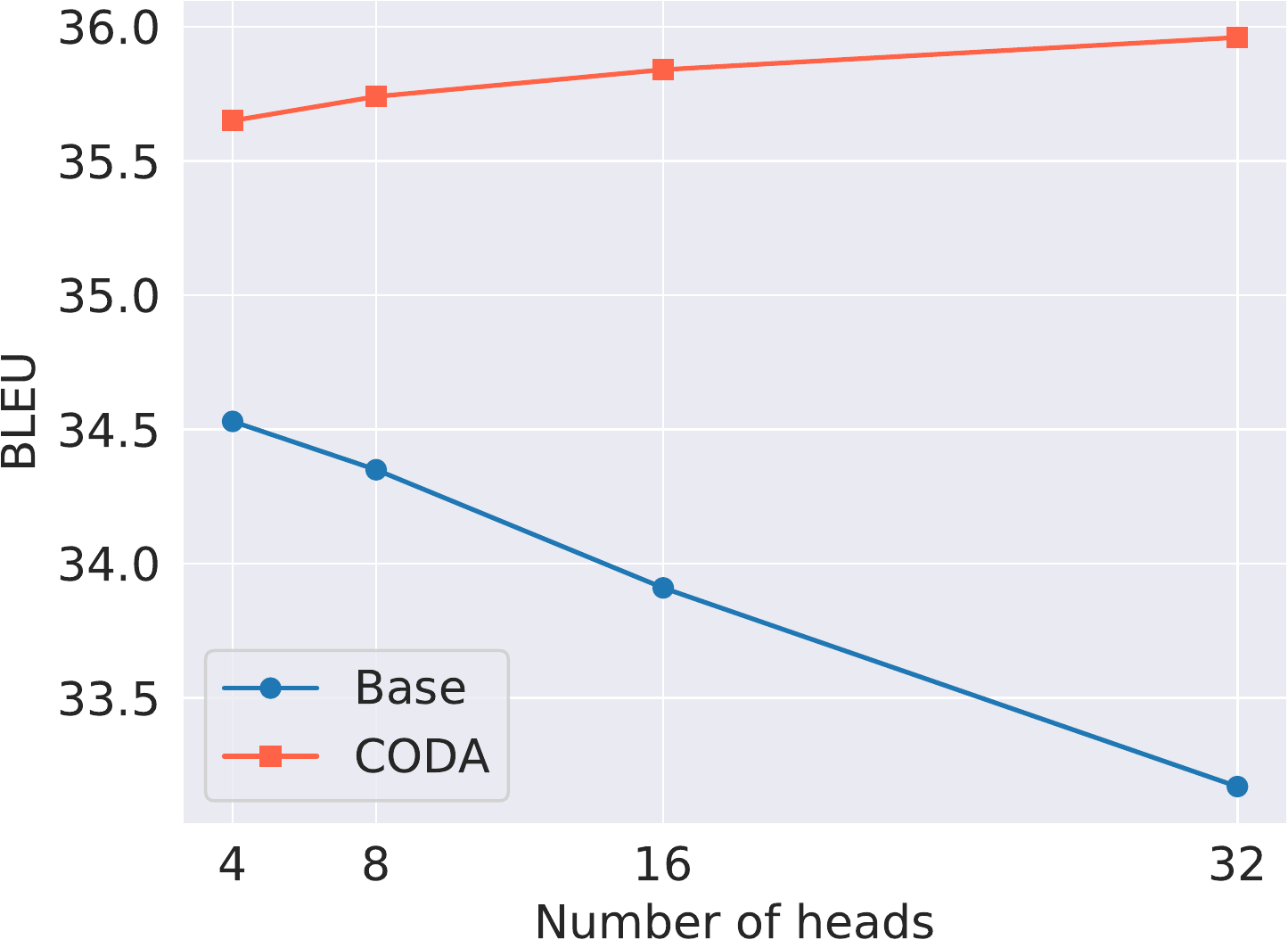}}
		\end{subfigure}
		\begin{subfigure}[b]{.49\columnwidth}
      \centering
      {\includegraphics[width=\columnwidth]{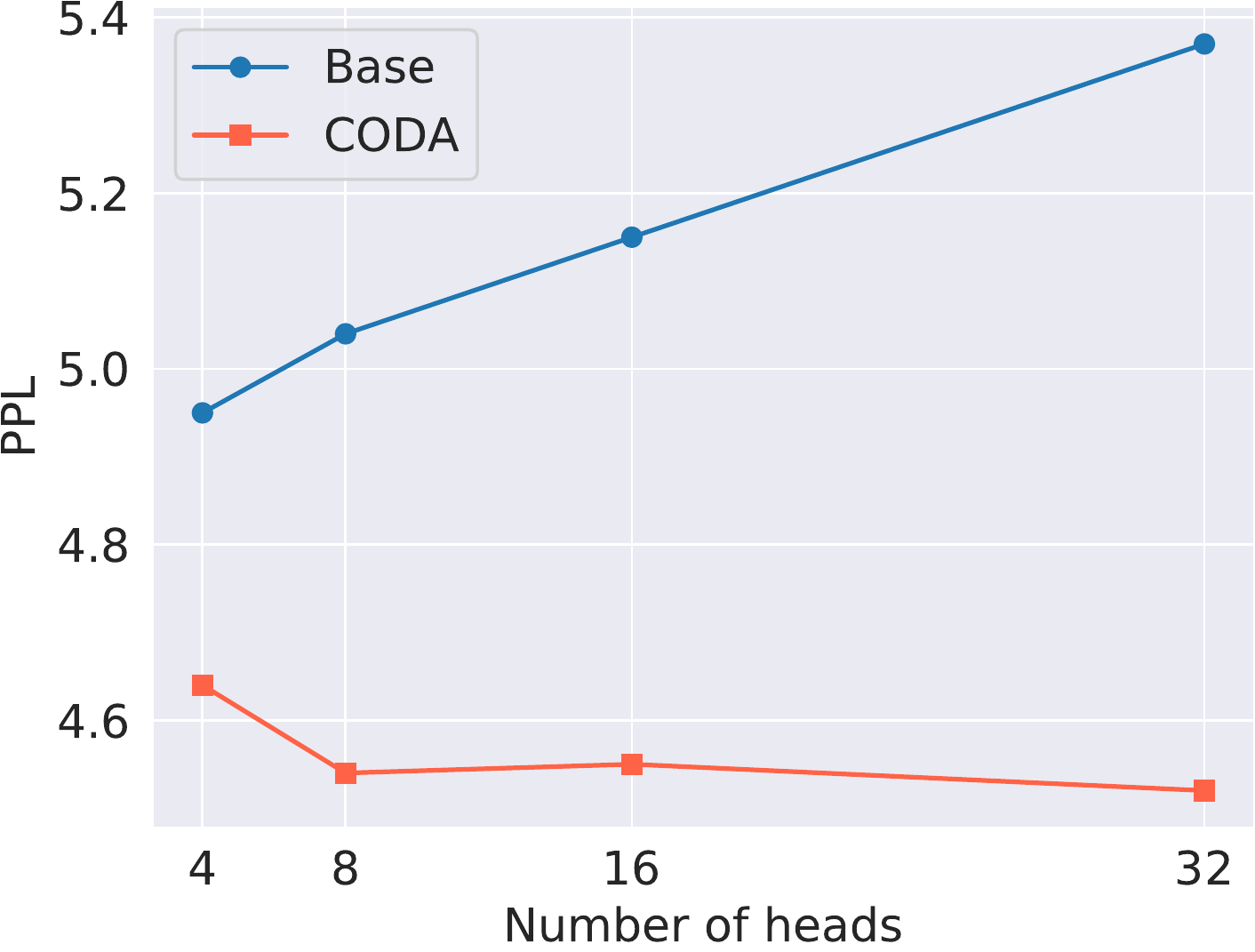}}
		\end{subfigure}
	\caption{Left: BLEU scores on test dataset for base transformers and \model under different number of attention heads (higher is better); Right: Perplexity on validation dataset for base transformers and \model under different number of attention heads (lower is better).}
	\label{fig:ablative-study-heads}
	
\end{figure}

Despite one would hope increasing the head number in MHA leads to a free-ride in achieving better performance, in practice it is often not the case as vanilla MHA suffers from the problem of parameter redundancy. Following \citet{vaswani2017attention}, we vary the number of attention heads (4,8,16,32), but keep the amount of computation constant. Our results on \iwslt are shown in Table~\ref{tb:ablative-study-heads}. We observe that the translation quality of baseline transformer (which uses vanilla MHA as its main building blocks) decreases almost linearly when increasing number of attention heads (Figure~\ref{fig:ablative-study-heads}), which agrees with previous studies~\citep{vaswani2017attention,voita-etal-2019-analyzing,michel-etal-2019-evaluation}.

Intuitively, since the total number of parameters in the model remains unchanged, more heads indicate that the number of parameters allocated to each head is reduced, which limits the representational power of every single attention head. Due to the independence assumption between the heads, many of them tend to focus on similar regions of the sequence, leading to a great waste of modeling capacity. 


\begin{table}
	\centering
	\resizebox{0.9\columnwidth}{!}{    
		\begin{tabular}{c |c c || c c}
			\toprule
			\multirow{2}{*}{\# heads} & \multicolumn{2}{c||}{BLEU} & \multicolumn{2}{c}{PPL} \\
			\cline{2-5}
               & \base & \model    & \base    &  \model     \\
			\hline
			4  & 34.53  & \textbf{35.65}  & 4.95 & \textbf{4.64}\\
			8  & 34.35  & \textbf{35.74}  & 5.04 & \textbf{4.54}\\
			16 & 33.91  & \textbf{35.84}  & 5.15 & \textbf{4.55}\\
			32 & 33.17  & \textbf{35.96}  & 5.37 & \textbf{4.52}\\
			\bottomrule
	\end{tabular}}  
	\caption{Left: BLEU scores on test dataset for \base and \model at different numbers of attention heads; Right: Perplexity on validation dataset for \base and \model at different numbers of attention heads.}
	\label{tb:ablative-study-heads}
\end{table}
In the case of \model, we observe better BLEU scores in response to the increasing head number. 
Rich interactions in \model could encourage different heads to cover broader regions of input sequence, which in turn offers more useful information for training. The perplexity (PPL) reflects similar trends. The coordination between different heads in \model greatly improves the model's parameter efficiency.

\subsection{Ablation analysis}
In this section, we present an ablation study to investigate effects of different components in \model. Concretely, we compare four models on the \iwslt machine translation task: (i) the full model \model, (ii) a variant of \model ablating the cascaded structure (\S\ref{sec:training}), (iii) a variant of \model without using head-colliding attention (\S\ref{sec:model}) and (iv) the baseline \base model.

In more details, for model (ii), we remove the second term in \eqref{eqn:direct-connection}, which turns off the direct cascading structure, despite still being a proper hierarchical latent variable model\footnote{Note that the first term $\widetilde{\mQ}_i\widetilde{\mK}_i^T$ in \eqref{eqn:direct-connection} also depends on the instantiated value of $\rvz_{i,j:}^{l-1}$, which induces an implicit hierarchical dependency for attention between adjacent layers.}.
In model \textbf{(iii)}, attention heads are deterministic (instead of being latent variables) as in vanilla Transformers, but cascading connections are incorporated. We observe its close connection with the recently proposed \textsc{realformer} \citep{he2020realformer}, a \base model that adds a residual connection between attention logits at adjacent layers. Since in model \textbf{(iii)} all attention heads are deterministic, it is unnecessary to fuse different heads (see \S\ref{sec:training}). In this case, we simply implement model (iii) as a \textsc{realformer} (and thus referred to as \textsc{realformer} hereafter) to demonstrate the effect of cascading-like structures more clearly.\footnote{The main difference between residual connections in \textsc{realformer} and cascading connections in \model is that, the former directly performs a \textit{head-wise} addition of previous-layer attention logits; in contrast, our cascading connection makes use of an MLP $\sigma(\cdot)$ to mix different attention heads, which enhances head interactions for \model.}

\begin{figure*}
\centering
\begin{subfigure}[b]{0.23\textwidth} 
  \centering
  {\includegraphics[width=\textwidth]{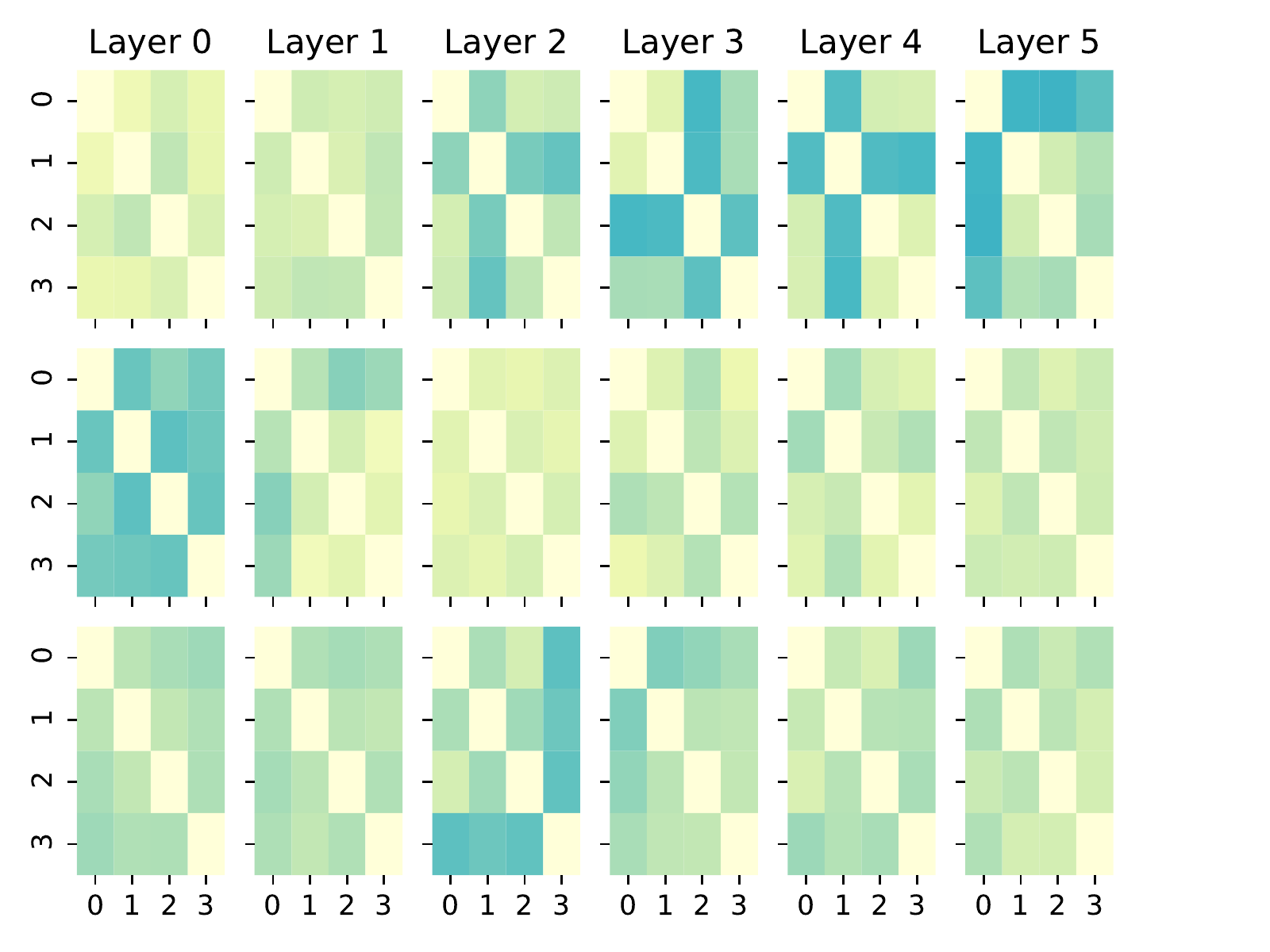}}
  \caption{\base}
\end{subfigure}
\begin{subfigure}[b]{0.23\textwidth} 
  \centering
  {\includegraphics[width=\textwidth]{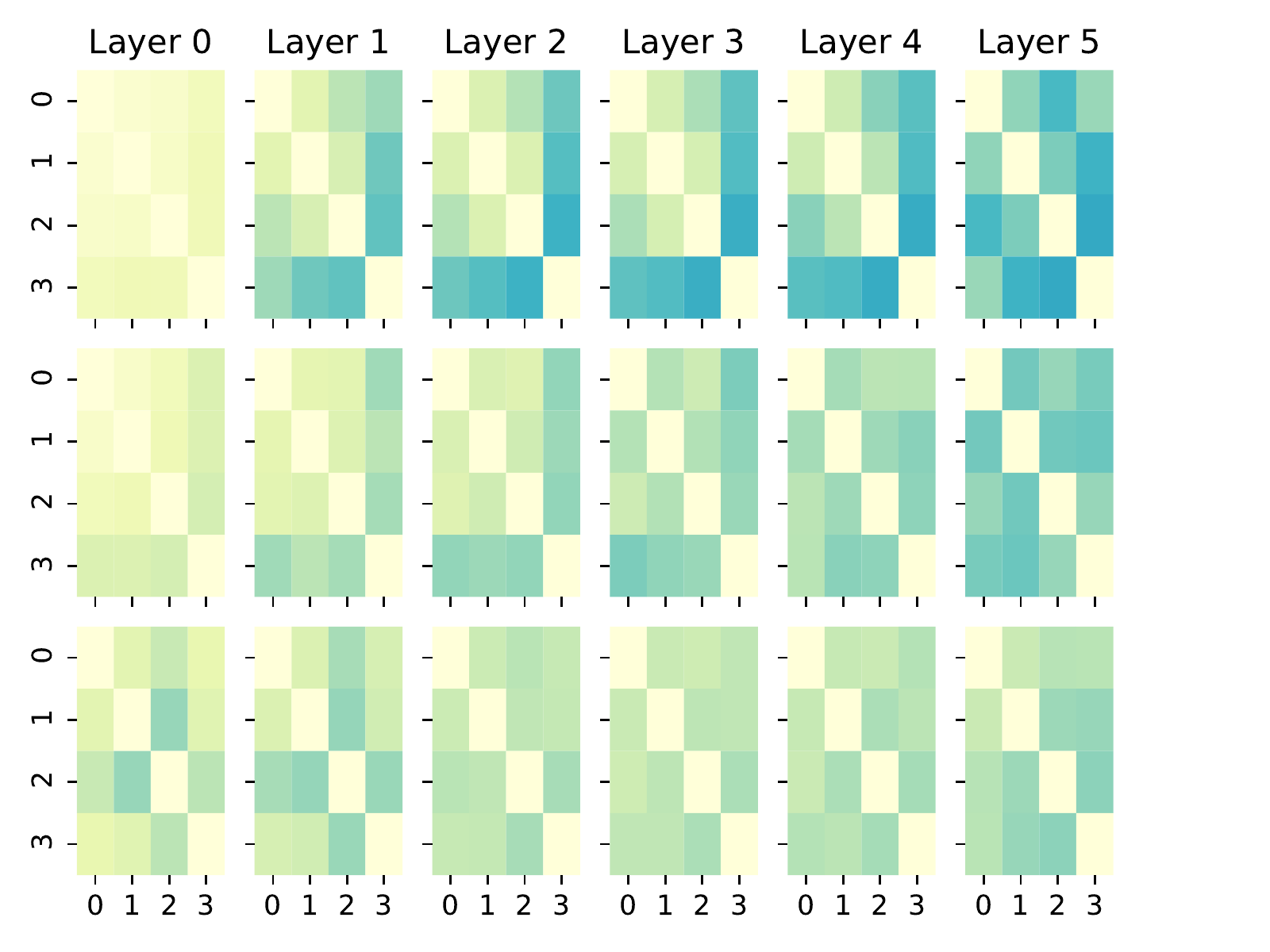}}
  \caption{\textsc{realformer}}
\end{subfigure}
\begin{subfigure}[b]{0.23\textwidth} 
  \centering
  {\includegraphics[width=\textwidth]{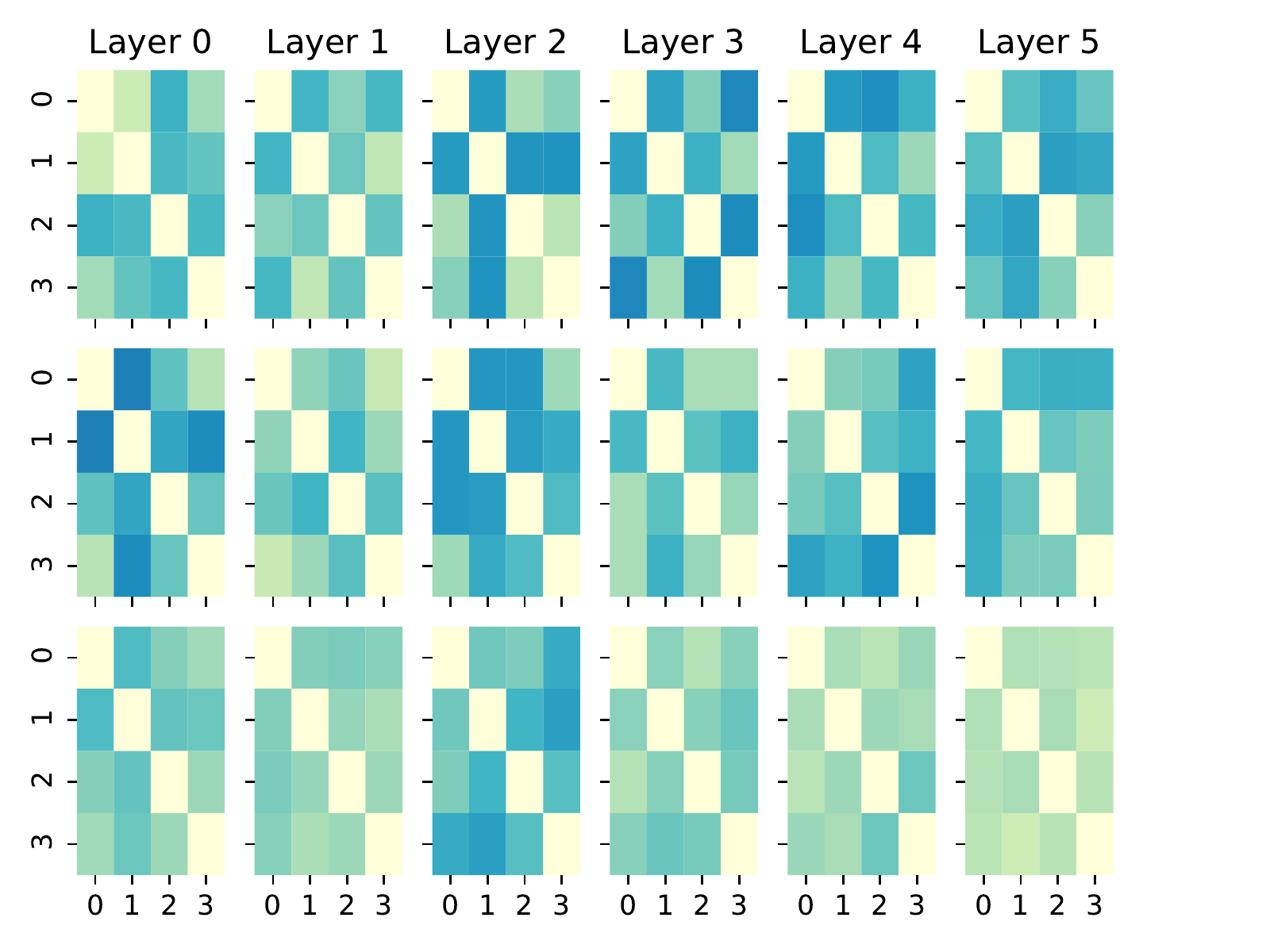}}
  \caption{\model -\textsc{cs}}
\end{subfigure}
\begin{subfigure}[b]{0.23\textwidth} 
  \centering
  {\includegraphics[width=\textwidth]{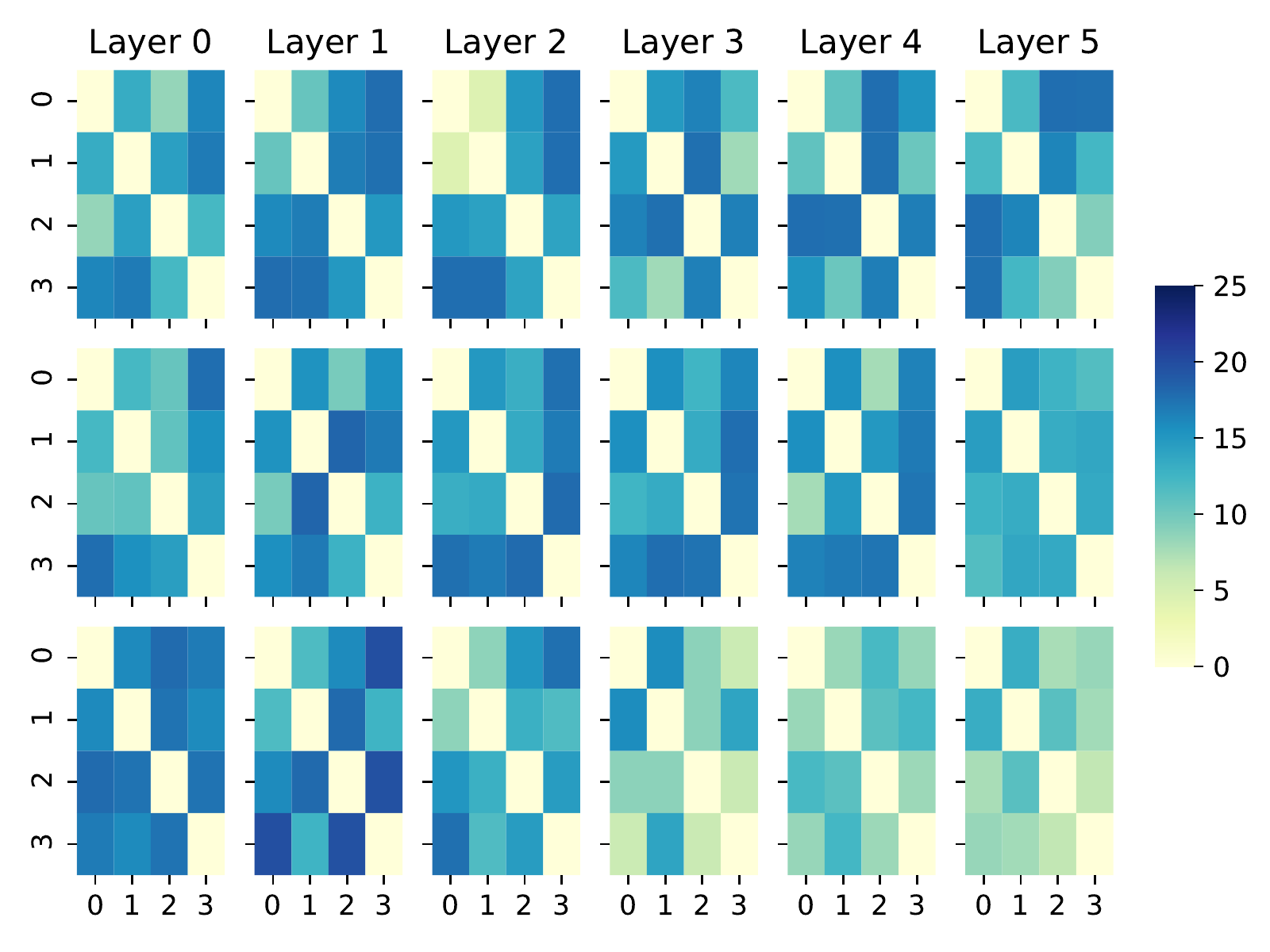}}
  \caption{\model}
\end{subfigure}
\caption{Jensen-Shannon Divergences (JSD) for each pair of attention heads at the same layer on \iwslt dataset for \base, \textsc{realformer}, \model -\textsc{cs} and \model model respectively. Each row indicates different kinds of attention, including encoder self-attention, decoder self-attention and decoder-encoder cross attention (from top to bottom), respectively; and each column indicates average JSD scores at different layers.}
\label{fig:jsd-abative}
\end{figure*}

We report BLEU score for translation quality, and the Jensen-Shannon Divergences (JSD) averaged over all heads pairs of all MHA blocks for quantitative evaluation of head interactions. As demonstrated in Table~\ref{tb:ablative-study-1} and Figure~\ref{fig:jsd-abative}, even without cascading connections for explicit hierarchical structures, head-colliding attention has the ability (albeit limited) to induce reasonable correlations among different heads, reflected in the average JSD. This is due to the explaining-away effects and the native hierarchical structure in the transformers, as discussed in \S\ref{sec:model}. In \model, because individual heads have access to the other heads from a probabilistic perspective, they are more prone to offering complementary information for each other to jointly explain the observed data. This effect is further enhanced when cascading connections are added to the model. 
In contrast, if we simply incorporate such cascading connections into a vanilla \base model, we found it does not significantly encourage head interactions and only improves the baseline marginally. In this case, the performance improvement might be mainly due to residual connections, which are often considered to be effective in facilitating training \citep{resnet}.
Interestingly, we note a positive correlation between average JSD and BLEU, suggesting that encouraging complementary attention heads may help improve translation quality.

\begin{table}
	\centering 
	\resizebox{0.8 \columnwidth}{!}{    
		\begin{tabular}{c||c|c}
			\toprule
			{Model}   & {Avg. JSD} & {BLEU} \\
			\hline
           \model    &  13.72 & 35.65     \\   
			\hline
			\model - \textsc{cs}  &  11.24  & 35.17     \\ 
			\textsc{realformer} \citep{he2020realformer} & 8.53 & 35.01\\
			 \base         &  7.11  & 34.53   \\
			\bottomrule
	\end{tabular}}
	\caption{The average JSD and BLEU scores with different model configurations. \model -\textsc{cs} indicates the ablation of the cascading structures from the full model (i.e., simply replacing all MHA blocks of base transformer with head-colliding attention); \textsc{realformer} is a recently proposed \base model that has cascading-like structures but still views each head as a deterministic value rather than latent variables.}
	\label{tb:ablative-study-1}
\end{table}

\section{Related Work}
Attention mechanisms were first applied to recurrent networks in \citep{bahdanau2014neural}. It was then extended to multi-head attention (MHA) and became the key component in transformer architectures \citep{vaswani2017attention}. 

To study the utility of multiple attention heads, \citet{voita-etal-2019-analyzing} focused on identifying individual contributions of each attention head. \citet{michel2019sixteen} conducted extensive experiments to demonstrate that pruning out most heads after training does not lead to a drop in performance during inference. \citet{you-etal-2020-hard} further revealed that replacing learnable attention heads with samples from fixed Gaussian distributions can achieve almost the same performance as original models. Additionally, \citet{behnke-heafield-2020-losing} proposed to iteratively prune attention heads during training based on the lottery ticket hypothesis. These works indicate that there is a lot of head redundancy in the MHA transformer architectures.

Instead of pruning unnecessary parameters and down-sizing transformer models, there are also works that propose to improve parameter efficiency in transformers. For instance, \citet{li-etal-2018-multi} introduced a regularization term to explicitly promote diversity among different heads. \citet{yang-etal-2019-convolutional} proposed to use convolutional kernels to capture correlations among not only local windows of sequences, but also different heads. \citet{an-etal-2020-repulsive} considered each head as a sample from the same distribution, and presented a sampling algorithm that avoids samples from collapsing into local modes. It hence explicitly encouraged the repulsiveness in MHA. Besides, MAE \citep{peng-etal-2020-mixture} converted a vanilla MHA to a mixture-of-experts model, where each expert component activates only a subset of attention heads. With learned probabilities, different experts could be specialized on different inputs. Different from these works, \model does not explicitly promote head diversity nor specialize different heads. Instead, we focus on studying head interactions from a probabilistic perspective, which reveals the close connection between vanilla MHA and \model.

Another research line relating to our work is to incorporate latent variables into attention modules. \citet{xu2015show} investigated the connection between vanilla deterministic single-head attention and its stochastic counterpart. \citet{deng2018latent} explored this further and proposed to use variational inference techniques for training the model. They considered both cases of discrete and continuous latent variables. Bayesian attention modules \citep{fan2020bayesian} introduced continuous latent distributions for attention that are amenable to reparameterization tricks. Our work is different from them in that we mainly investigate the MHA mechanism and aim to improve parameter-efficiency by recovering potential interactions among different heads, which are ignored in vanilla MHA.


Concurrently, \citet{he2020realformer} proposed to add residual connections between attention scores at adjacent layers, similar to our cascading connections. Nevertheless, our motivation for using the cascaded structure is quite different: we aim to construct direct hierarchical dependencies for latent variable models, while \citet{he2020realformer} is mainly motivated to improve transformer architectures and obtain performance gains.

\section{Conclusion and Future Work}
We present \model by re-formulating the multi-head attention (MHA) as a latent variable model from a probabilistic perspective. \model explicit models of the interactions among attention heads through a hierarchical variational distribution. We conduct extensive experiments and demonstrate that \model outperforms the transformer baseline in language modeling and machine translation. The analysis shows that \model learns to encourage the diversity in different heads and to promote parameter efficiency when increasing the number of heads. In this framework, we will be able to impose explicit constraints or regularization on different attention heads in a principal way (e.g. informative priors that promote diversity). Besides, we can also consider more expressive (data-driven) variational distributions. We leave these as the future work. Our code is publicly available at \url{https://github.com/LZhengisme/CODA}.


\section*{Acknowledgments}
We thank the anonymous reviewers whose suggestions helped clarify this work. This research was supported in part by the University of Hong Kong Research Committee under account 104006039.111994.14200.301.01.

\bibliographystyle{acl_natbib}
\bibliography{anthology,acl2021}

\clearpage
\appendix
\section{Implementation details}
\label{app:implementation}
For the $\sigma$ network, it consists of a 2-layer MLP with \texttt{LeakyRelu} non-linear activation and a residual link from the input. It is a rather small network and only accounts for 0.01-0.02$\%$ of the total parameters. Recall that the number of attention heads is denoted by $h$, the source and target length is $m$ and $n$ respectively, and the batch size is denoted by $b$. The hidden size is set to $\alpha * h$, where we select $\alpha$ from $\{2,4,8\}$ based on the validation set. Note that the additionally introduced number of parameters is negligible compared to the model size, accounting for only 0.01-0.02\% of the total parameters. Since we often represent the attention scores (or logits) $\rvz$ as a multi-dimensional tensor with shape \texttt{(b, h, n, m)}, we first transpose it to shape \texttt{(b, m, n, h)} and feed it into the $\sigma$ network. It then outputs $h$ values so that each component $\sigma_i$ computes the fused information from all previous layer's attention heads. By adding its output to the current layer's attention logits, we could effectively construct a direct cascading connection for our hierarchical proposal. Note that $\sigma$ network is neither shared among different heads nor different layers.
\subsection{Machine translation}
For \wmt, the transformer-base architecture in \citet{vaswani2017attention} is used, where both the encoder and decoder consist of 6 layers with hidden size 512. For MHA blocks at each layer, the number of attention heads is set to 8 with the dimension of hidden layer representations being 512; For feed forward networks, the hidden size is set to 2048. The rate of dropout is set to 0.1. For training, we follow the same setup as in \citet{vaswani2017attention}, including that label smoothing with rate 0.1, the Adam optimizer \citep{kingma2014adam} is used for optimization, the inverse square root scheduling is utilized for learning rate and the number of warm-up steps is set to 4000. 

For IWSLT-14, we follow the configuration of hyper-parameters in Fairseq package \footnote{\url{https://github.com/pytorch/fairseq/tree/master/examples/translation}}. In details, it mostly follows the same architecture and training setup as above, except that it uses a smaller feed forward network with hidden dimension 1024, a larger dropout rate 0.3 and less attention heads 4.

For both datasets, we apply a compound split post-processing to facilitate comparison. Additionally, we use activation dropout with rate 0.1 for all used models on both datasets as we find it helps our model converge better.

\subsection{Language modeling}
For \texttt{Wikitext-103}, we base our model on \citet{baevski2018adaptive} with the same hyper-parameter configuration and training setup. The model architecture consists of 16 transformer layers, where it uses adaptive input representations, 8 heads for each MHA block, dropout rate of 0.3, hidden dimension of 1024, and hidden size of 4096 for feed forward networks. For training, Nesterov’s accelerated gradient (NAG) method \citep{nag2013} is used with gradient norm clipping and a cosine learning rate schedule\footnote{More details can be found in \citet{baevski2018adaptive} and the training script based on Fairseq codebase: \url{https://github.com/pytorch/fairseq/blob/master/examples/language_model/README.adaptive_inputs.md}.}.

\section{Additional experimental results}
\label{app:figs}
\begin{figure*}
\centering
\begin{subfigure}[b]{0.4\textwidth} 
  \centering
  {\includegraphics[width=\textwidth]{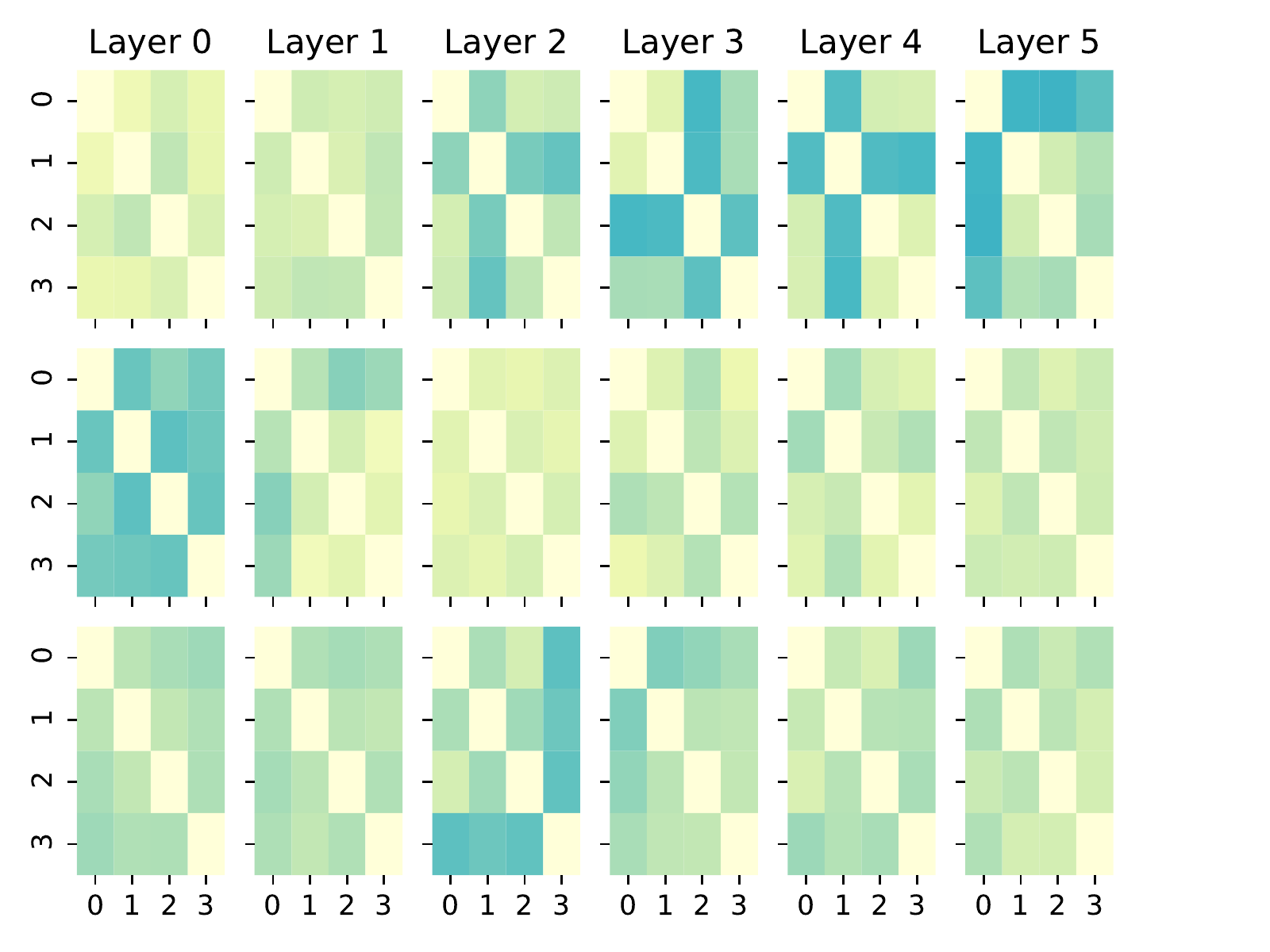}}
  \caption{\base}
\end{subfigure}
\begin{subfigure}[b]{0.4\textwidth} 
  \centering
  {\includegraphics[width=\textwidth]{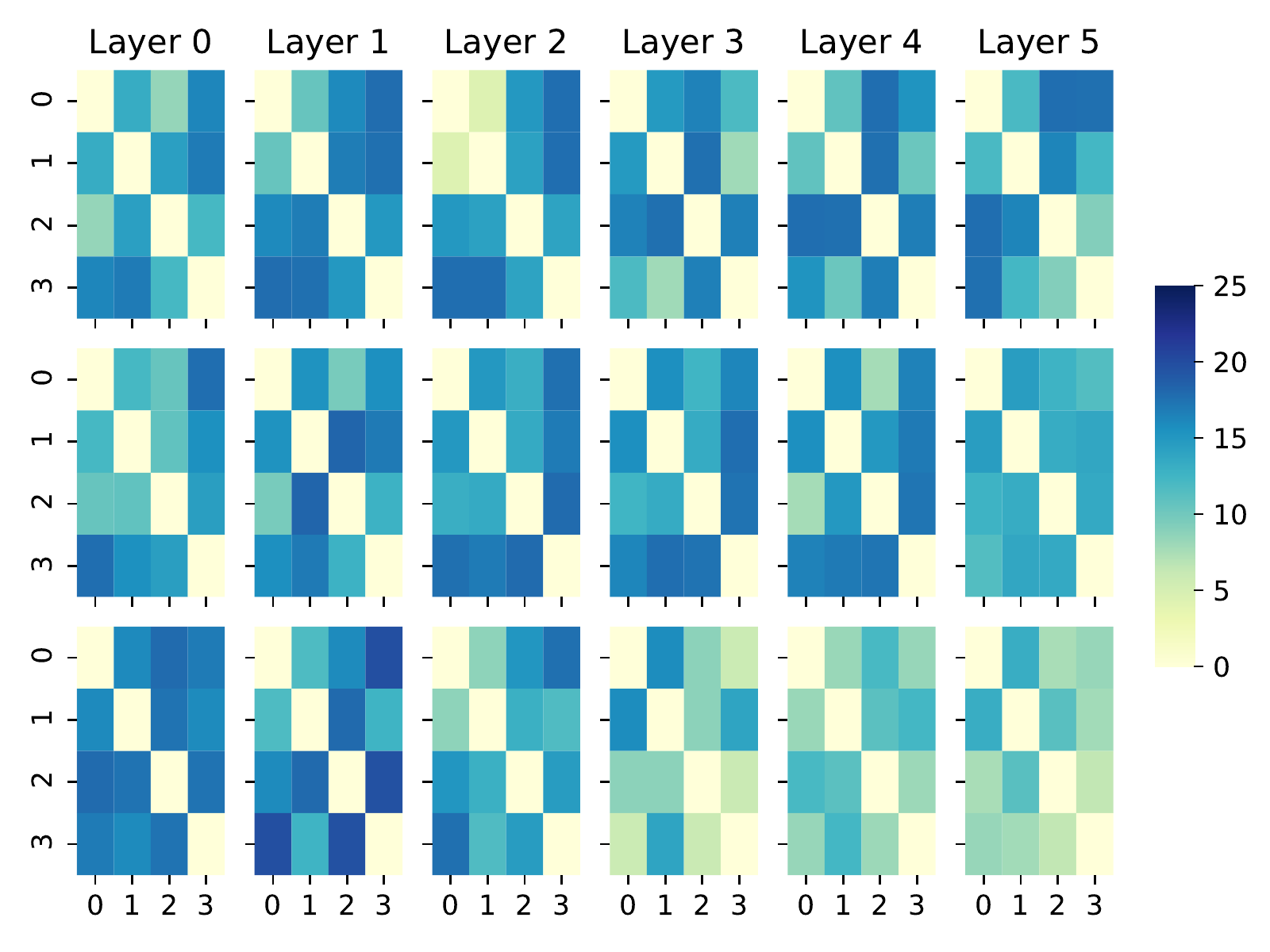}}
  \caption{\model}
\end{subfigure}
\caption{Jensen-Shannon Divergences (JSD) for each pair of attention heads at the same layer on \iwslt validation dataset, which are evaluated on both \base model and \model. Each row indicates different kinds of attention, including encoder self-attention, decoder self-attention and decoder-encoder cross attention (from top to bottom), respectively; and each column indicates average JSD scores at different layers.}
\label{fig:jsd-iwslt-4heads}
\end{figure*}

\begin{figure*}
\centering
\begin{subfigure}[b]{0.42\textwidth} 
  \centering
  {\includegraphics[width=\textwidth]{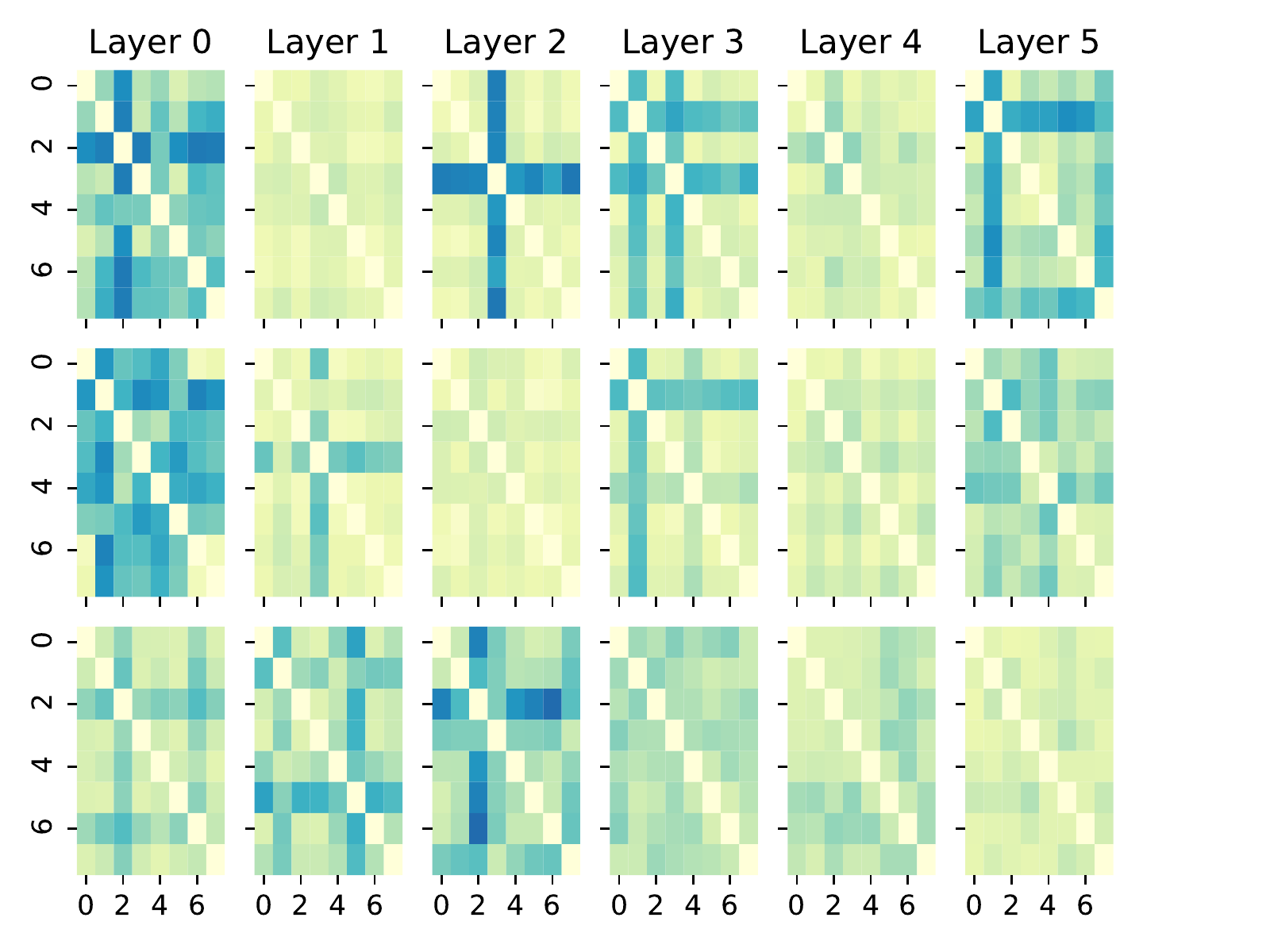}}
  \caption{\base}
\end{subfigure}
\begin{subfigure}[b]{0.42\textwidth} 
  \centering
  {\includegraphics[width=\textwidth]{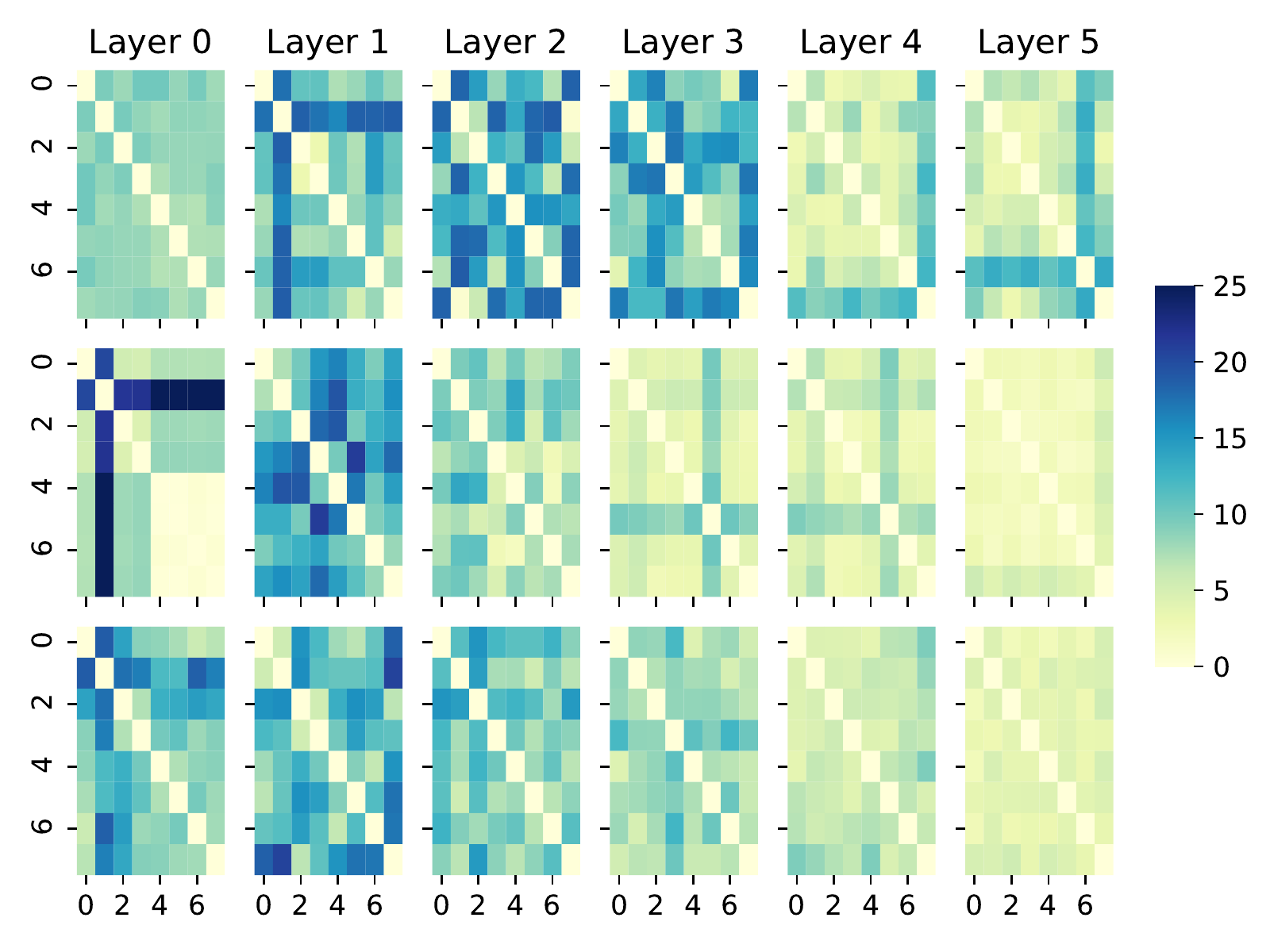}}
  \caption{\model}
\end{subfigure}
\caption{Jensen-Shannon Divergences (JSD) for each pair of attention heads at the same layer on \wmt dataset, which are evaluated on both \base model and \model. Each row indicates different kinds of attention, including encoder self-attention, decoder self-attention and decoder-encoder cross attention (from top to bottom), respectively; and each column indicates average JSD scores at different layers.}
\label{fig:jsd-wmt-8heads}
\end{figure*}

Figure~\ref{fig:jsd-iwslt-4heads} and Figure~\ref{fig:jsd-wmt-8heads} visualize head interactions within \base and \model on \iwslt and \wmt translation tasks respectively. 

\end{document}